\pdfoutput=1

\documentclass[11pt]{article}

\usepackage[final]{acl}

\usepackage{times}
\usepackage{latexsym}
\usepackage{float}
\usepackage{array}
\usepackage{subcaption} 
\usepackage[T1]{fontenc}
\usepackage[utf8]{inputenc}
\usepackage{microtype}
\usepackage{inconsolata}
\usepackage{graphicx}
\usepackage{tabularx}
\usepackage{booktabs}
\usepackage{caption}
\usepackage{subcaption}
\usepackage{colortbl}
\usepackage[dvipsnames]{xcolor}
\usepackage{chngcntr}
\usepackage{apptools}

\title{WildFireCan-MMD: A Multimodal Dataset for Classification of User-Generated Content During Wildfires in Canada}

\author{
    Braeden Sherritt\textsuperscript{\rm 1}, Isar Nejadgholi\textsuperscript{\rm 2}, Efstratios Aivaliotis\textsuperscript{\rm 1},\\{\bf Khaled Mslmani\textsuperscript{\rm 1}, Marzieh Amini\textsuperscript{\rm 1}}\\
    \textsuperscript{\rm 1}Carleton University, Ottawa, Canada\\
    \textsuperscript{\rm 2}National Research Council Canada, Ottawa, Canada\\
    \small \texttt{BraedenSherritt@cmail.carleton.ca,isar.nejadgholi@nrc-cnrc.gc.ca,Efstratiosaivaliotis@cmail.carleton.ca,}\\
    \small \texttt{KhaledMslmani@cmail.carleton.ca, MarziehAmini@cunet.carleton.ca}
}

\begin{document}
\maketitle
\begin{abstract}
Rapid information access is vital during wildfires, yet traditional data sources are slow and
costly. Social media offers real-time updates, but extracting relevant insights remains a challenge. In this work, we focus on multimodal wildfire social media data, which, although existing in current datasets, is currently underrepresented in Canadian contexts. We present WildFireCan-MMD, a new multimodal dataset of posts on X from recent Canadian wildfires, annotated across twelve key themes. We evaluate zero-shot vision-language models on this dataset and compare their results with those of custom-trained and baseline classifiers. We show that while baseline methods and zero-shot prompting offer quick deployment, custom-trained models outperform them when labelled data is available. Our best-performing custom model reaches 84.48±0.69\% f-score, outperforming VLMs and baseline classifiers. We also demonstrate how this model can be used to uncover trends during wildfires, through the collection and analysis of a large unlabeled dataset. Our dataset facilitates future research in wildfire response, and our findings highlight the importance of tailored datasets and task-specific training. Importantly, such datasets should be localized, as disaster response requirements vary across regions and contexts.

\end{abstract}

\section{Introduction}
\label{sec:introduction}

Large-scale analysis of online interactions, using machine learning algorithms has enabled critical applications such as disaster response \cite{Fauzi2023,kumar2024extracting}, mental health monitoring \cite{ghanadian2024socially,dalal2024review}, detecting and combating hate speech in digital environments \cite{kiritchenko2021confronting,kumar2024detecting}, pandemic discourse analysis \cite{tbcov19} and sentiment and behavioral trend analysis \cite{ma2024toward,curto2024crime}. Among these applications, analysis of social media data during natural disasters has gained increasing importance in recent years \cite{Alam2018, Ofli2020, alam2021humaid, Dwarakanath2021, Algiriyage2022, Hossain2022, Khattar2022, Basit2023, Koshy2023, Rezk2023}. Such data contains crucial details such as reports of casualties, infrastructure damage, and urgent needs of affected individuals, among other critical information. The effective use of the shared information can facilitate humanitarian aid efforts for more efficient management and faster recovery \cite{Houston2015}. This valuable data, however, is mixed in with noise, such as humorous content, advertisements, and other irrelevant information. Therefore, there is a critical need to develop an automated method for sifting through this information. 

To foster research in this area, \citet{Alam2018, alam2021humaid} developed two multimodal datasets, CrisisMMD and HumAID, both collected from Twitter, including several disasters such as earthquakes and hurricanes. However, the classes of these datasets are generalized across all types of disasters, meaning they are not tailored to specific disaster scenarios. As a result, the classification scheme is broad, and certain nuances or unique information that may be relevant to a particular type of disaster could be overlooked. Also, while the CrisisMMD dataset does contain wildfire data from the California wildfires of 2017, the proportion of this type of data compared to the rest is relatively small, only comprising 1589 samples, and is specific to the California wildfires of October 2017. The wildfire-related samples from Canada in the HumAID dataset \cite{alam2021humaid} are limited to a single month in 2016, as well as another month of California samples from 2018. Therefore, wildfires in Canada remain underexplored in social media analysis research.

To address this limitation, we focus our work on gathering social media data specifically related to Canadian wildfires, as they are becoming increasingly frequent and severe due to the adverse effects of climate change. Upon initial inspection of our Canadian wildfire-specific social media dataset, we observed that existing categorization schemes do not capture the specificity of this data. To address this, we devised a novel taxonomy of twelve categories reflecting the unique information needs and communication patterns of Canadian wildfire contexts. Using our developed taxonomy, we manually labeled 4,688 wildfire-related tweets from Canada and created \textbf{WildFireCan-MMD}, a new multimodal dataset of X posts from recent Canadian wildfires.

To benchmark the available classification models on WildFireCan-MMD, we evaluated the zero-shot classification capabilities of several Vision-Language Models (VLMs), custom deep-learning classifiers, as well as classical approaches, as a minimum achievable performance. Our best-performing classifier utilizes RoBERTa \cite{roberta2019} and Vision Transformer (ViT) models to jointly process the text and image components of tweets. A custom transformer-based fusion module is designed to enable the modalities to interact deeply. While WildFireCan-MMD is sourced from data posted between 2022 and 2024, we collected an additional set of unlabeled tweets spanning seven years (2018-2024). Using our best-performing classifier, we analyze this unlabeled data and uncover trends and patterns of social media use during wildfires in Canada. 
The full set of labeled data (containing the tweet IDs and labels), as well as the experimental code, is available on Github (\url{https://github.com/Multimodal-Social-Media-Data-Analysis/WildfireCanMMD}). The contributions of this study are summarized as:

\begin{itemize}
\item We collected 4,688 wildfire-related posts from X in Canada from 2022 to 2024 and devised a twelve-category taxonomy using unsupervised topic modeling and human analysis. We introduce \textbf{WildFireCan-MMD}, the first Canada-specific multimodal wildfire social-media dataset, annotated by three annotators.
\item We propose a custom multimodal transformer model, trained on WildFireCan-MMD and compare it with several VLMs and classical methods. The fine-tuned custom classifier outperforms zero-shot VLMs significantly; our model achieves an f-Score of 84.48±0.69\%.
\item Additionally, we collected 46,279 posts from X from 2018 to 2024. We used our trained classifier to label them and reported the trends and patterns of social media use during wildfires in Canada. Results show that retrospective analysis using a classifier can uncover trends which relate to real situational updates.
\end{itemize}

\section{Background and Literature Review}

\subsection{Social media analysis for disaster response }

Significant research has been conducted on the role of social media during disasters. \citet{Fauzi2023} conducted a bibliometric analysis of social media use in disaster response and offered a comprehensive overview of key studies and emerging research directions. \citet{Acikara2023} highlighted the potential of social media analytics in disaster response by leveraging users as ``citizen sensors'' to rapidly gather critical, time-sensitive information. \citet{Muniz2020} focused on the public health aspects of social media use during emergencies, noting its effectiveness in spreading emergency warnings and assessing needs after a disaster. In other works, \citet{Dwarakanath2021}, \citet{9220091}, and \citet{10485423} discussed the crucial role of social media in managing crises, pointing out how platforms like Facebook and Twitter support communication and coordination among affected individuals and communities. Furthermore, in a systematic review of the application of deep learning in disaster response, \citet{Algiriyage2022} discussed the successes and challenges faced in this area, proposed guidelines for future research, and highlighted the importance of using multimodal data in disaster scenarios. Our work is motivated by this literature as well as the literature shown in Appendix \ref{appendix:litreview} which focuses this work on using machine learning models to triage Canada-specific multimodal social media data to inform disaster response and management. 

\subsection{Multimodal classification for disaster response}

We review this literature based on the available datasets and previously explored algorithms.

\vspace{5pt}
\noindent \textbf{Datasets:} 
While many works utilize textual datasets sourced from social media to classify posts during disasters \cite{crisislex2014, Dong2021, Ochoa2021}, it is now generally considered that multimodal approaches offer stronger results \cite{Ofli2020, Abavisani2020, ramirez2025}.

The study by \citet{Alam2018} released CrisisMMD, a collection of tweets (~18k samples) with images from seven major natural disasters (including wildfires, earthquakes, floods, and hurricanes). CrisisMMD contains thousands of tweets labeled for categories like “infrastructure and utility damage” or “rescue volunteering or donation effort”.

Additionally, \citet{alam2021humaid} later released the HumAID dataset which consists of ~77K human-labeled disaster tweets sampled from ~24M tweets over 19 events (2016–2019). It is similarly annotated for emergency response categories (e.g. casualties, infrastructure damage).

There are several other multimodal datasets that exist, but most are variations or combinations of the CrisisLex \cite{crisislext262015} and CrisisMMD datasets \cite{Xuehua2022}. CrisisMMD has become a foundational resource for studying disaster-related communication patterns and the combined use of text and images. It remains the most prominent dataset in this domain, used in many recent studies \cite{Ofli2020, Khattar2022, Basit2023, Koshy2023, Rezk2023} and many others. Therefore, there is a notable gap in the in the availability of multimodal wildfire datasets. 

In the context of Canada, where the most common and destructive natural disaster is wildfires, CrisisMMD is lacking in data. Although it contains ~18k samples from other disasters such as hurricanes, earthquakes and floods, the amount of wildfire data is proportionally small (1589 samples). Additionally, this data is only from the California wildfires of October 2017. The wildfire-related samples from Canada in the HumAID dataset come from only a single month in 2016 (2,259 samples), as well as another month of California samples from 2018 (7,444 samples). Therefore, wildfires in Canada remain underexplored in social media analysis research. To address this limitation, this work focuses on gathering social media data related to Canadian wildfires, and proposes WildFireCan-MMD, a new Canadian wildfire social-media dataset with a wildfire-specific taxonomy 12 categories.

\vspace{5pt}
\noindent \textbf{Algorithms:} Previous work by \citet{Ofli2020} introduced a joint representation from two parallel deep learning architectures where one represents the text modality and the other represents the image modality. Image and text modalities are vectorized through VGG16 and Word2Vec+CNN architectures respectively, followed by a dense layer for prediction. The work by \citet{Basit2023}, expands on \citet{Ofli2020} by experimenting with multiple recent transformer structures, for both the text and image modalities, and compared them against the baseline reported in the previous work. They found that using a combination of AlBERT \cite{albert2020} as a text embedder, and DeIT \cite{deit2021} as an image embedder, provided the best results for the eight-class classification task. \citet{Koshy2023} implemented further experiments with text and image transformers and showed improvements by using bi-LSTMs instead of LSTMs. Furthermore, they have noted that multiplicative fusion strategies produce better results on a classification task when combined with the previously described methods. Specifically, they demonstrated that using the fine-tuned RoBERTa \cite{roberta2019} model for text, Vision Transformer model for image, biLSTM+attention mechanism with multiplicative fusion, for prediction, achieves the best results for the eight-class classification task described by \citet{Alam2018}.

Many architectures have been proposed in other works for classification on CrisisMMD, as seen in Appendix \ref{appendix:litreview}. Most previous studies use a CNN for the image modality and then a BERT-based transformer for the text modality. Other works sometimes use ViT for the image modality. Concatenative fusion of the two modalities into a series of linear layers seems to be the most common approach, with other works using multiplicative or additive fusion. Cross-attention fusion of the two modalities seems to be another popular approach that seems to give promising results. The approach given by \citet{pranesh2022} proposes a transformer-like structure to fuse the two modalities. We build on the this previous work and explore the existing structures on the newly annotated data, WildFireCan-MMD. 

In addition to fine-tuned structures, we explore using VLMs in a zero-shot setting \cite{Duzhen2024, Li2022}. Specifically, we evaluate four VLMs, GPT-4o-mini \cite{OpenAI2023}, LLaVA-v1.5-13B \cite{Liu2023}, Qwen2.5-VL-7B \cite{qwen2025}, and SmolVLM-2B \cite{smolvlm2025}, to assess performance across a range of capabilities and computational requirements. GPT-4o-mini serves as a state-of-the-art, closed-source benchmark, while LLaVA and SmolVLM are open-source models that use visual instruction tuning and are optimized for varying levels of hardware, reflecting ongoing efforts to build accessible, high-performing multimodal systems.

\section{The WildFireCan-MMD Dataset}

\subsection{Data Collection}
\label{subsec:collection}
To construct the dataset, X’s Pro-Tier paid API was utilized to collect a diverse set of posts, including images and text, surrounding the British Columbia and Alberta wildfires of 2022 and 2023, as well as the Jasper wildfires of 2024. The dataset comprises 4,688 unique image-text pairs gathered from the wildfire season (April-September) of each respective year. Data collection was conducted using a hashtag-based search strategy as seen in other works \cite{Alam2018, gupta2020, Cruickshank2020}. Table \ref{tab:queries} presents the full queries used for data collection. To curate the hashtags, basic hashtags such as "\#wildfire, \#forestfire" were first used. Then, through an iterative process of adding new hashtags by collecting sample tweets, more relevant hashtags were discovered. This approach of using hashtags was necessary due to the limited availability of geotagged content, as most users on the platform do not enable precise geo-location. Instead, hashtags identified through iterative experimentation were found to be effective indicators of relevant content originating from Canada during wildfire events. The specificity of these hashtags ensured that the collected tweets were not only on-topic but also predominantly from users in the targeted geographic regions. The dataset includes the timestamp of when the tweet was posted, the user-provided location, and, in some cases, the specific region from which the tweet originated.

\begin{table}[h]
\small
    \centering
    \begin{tabular}{p{1.5cm}|p{5cm}}
    \hline
    \textbf{Year} & \textbf{Query}\\
    \hline
      2022/2023  & (\#BCwildfire OR \#BCfire OR \#ABWildfire OR \#albertawildfire OR \#ABFire) -has:videos has:images lang:en -is:retweet -is:quote -is:reply\\
      \hline
      2024  & (\#BCwildfire OR \#BCfire OR \#ABWildfire OR \#albertawildfire OR \#ABFire OR \#JasperStrong OR \#JasperWildfire OR \#JasperAB) -has:videos has:images lang:en -is:retweet -is:quote -is:reply\\
      \hline
    \end{tabular}
    \caption{The queries used for collecting the data for WildFireCan-MMD.}
    \label{tab:queries}
\end{table}

\begin{table*}[h] 
\centering
\scriptsize 
\renewcommand{\arraystretch}{1.3} 
\setlength{\tabcolsep}{5pt} 
\caption{Categories of WildFireCan-MMD, their descriptions and total count of each class. }
\label{tab:cat_explained}
\begin{tabular}{l p{10cm} c c} 
\hline
\rowcolor[gray]{0.8} 
\textbf{Category} & \textbf{Description} & \textbf{Count} \\
\hline
\textbf{Evacuees} & Information relating to evacuees, their movements, needs, location. & 252\\
\rowcolor[gray]{0.95} 
\textbf{Smoke and Air Quality} & Tweets related to or showing signs of smoke or the current air quality. & 1128\\
\textbf{General Information} & General info about the wildfire situation, such as total hectares burned. & 170\\
\rowcolor[gray]{0.95} 
\textbf{Preparedness} & Information for the general public to prepare themselves and their property. & 264\\
\textbf{Weather Reports} & Information relating to the weather with a specific location mentioned. & 296\\
\rowcolor[gray]{0.95} 
\textbf{Warnings \& Status Updates} & Fire bans in certain areas, new information about a specific area, updates from officials. & 669\\
\textbf{Reports of Actions of Responders} & Actions of responders within specific areas or times, including prescribed burns. & 356\\
\rowcolor[gray]{0.95} 
\textbf{Infrastructure} & Detours, road closures, damage to infrastructure (e.g., utility poles, highways), repairs by crews. & 264\\
\textbf{Political} & Posts directed towards political figures or parties (excluding situation updates). & 329\\
\rowcolor[gray]{0.95} 
\textbf{Support} & Mental health and financial support, temporary housing for livestock. & 178\\
\textbf{Insurance} & Information relating to insurance, employment insurance, and EI benefits. & 158\\
\rowcolor[gray]{0.95} 
\textbf{Advertisement} & Posts about food, restaurants, off-topic ads for services (e.g., apps, air purifiers, not insurance-related). & 117\\
\textbf{Other} & No 'useful' information, focusing on images (e.g., scenery), general complaining, or irrelevant content. & 507\\
\hline
\end{tabular}
\end{table*} 

\subsection{Taxonomy Development}
\label{subsec:clustering}

To examine the content of the tweets and discover the hidden latent semantics of the data, BERTopic was employed to perform multimodal unsupervised topic modeling \cite{Grootendorst2022}. Text and image inputs were vectorized using the pre-trained ``Clip-ViT-B-32''\footnote{\url{https://huggingface.co/sentence-transformers/clip-ViT-B-32}} model as suggested by BERTopic. The default parameter settings of BERTopic were used, which reduced the dimensionality to 5, while keeping the size of the local neighborhood at 15. Stop words were removed, and both unigrams and bigrams were considered as candidates of topic keywords. Using BERTopic on the dataset, the algorithm determined the number of topics to categorize the data into, resulting in 100 fine-grained topics. Hierarchical topic reduction was applied using Top2Vec, and the number of topics was gradually decreased. For each iteration, the representative text and visual samples of topics and keywords generated by BERTopic were qualitatively inspected. Once the number of topics was reduced to around 15, it became clear which topics were the strongest in the dataset, and the humanitarian utility of some of these topics began to appear. Some examples of useful topics included posts about infrastructure, weather reports, and reports on the actions of first responders. Less useful posts were also noticed, such as advertisements and political debates/complaints. Some examples of these topics, along with representative images and corresponding keywords generated by BERTopic, are presented in Appendix \ref{appendix:bertopic}. 

The topics were inspected and 12 classes were decided upon. The full list of classes and their descriptions is given in Table \ref{tab:cat_explained}. To ensure the reliability of the labeling process, the definitions and boundaries of each category were discussed among the research team and a shared understanding was reached. Through this discussion, it was determined that the classes are well-defined and grounded in observable cues (e.g., specific topics, hazards, or references in the text or image). This discussion, along with unsupervised topic modeling (BERTopic) were used to find strong topics which were then later assigned to a class. Also, the selection of classes was guided by the topics created by BERTopic, observations made through manual annotation, and the perceived relevance of classes to stakeholders in disaster scenarios. For example, \textit{‘Warnings \& Status Updates’}, \textit{‘Reports of Actions of Responders’}, and \textit{‘Infrastructure’} could be useful to first responders and emergency personnel. The \textit{‘Evacuees,’} ‘\textit{General Information,’} and \textit{‘Preparedness’} categories could be useful to the general public to be informed about the events happening around their locality. The \textit{‘Political’} category could be useful to the media to gauge public opinion and policies on events surrounding the disaster. An \textit{`Other'} class is added to accommodate the samples that do not fit into any of the 12 classes.

\begin{table}[ht]
    \centering
    \caption{Annotation agreement statistics.}
    \small
    \begin{tabular}{p{5cm}|c}
    \hline
    \textbf{Metric} & \textbf{Agreement}\\
    \hline
    Majority Agreement (2 same) & 88.4\%\\
    \hline
    Full Agreement (all same) & 45.1\%\\
    \hline
    Vote between all/annotator 1 & 78.7\%\\
    \hline
    Vote between all/annotator 2 & 77.7\%\\
    \hline
    Vote between all/annotator 3 & 65.8\%\\
    \hline
    Cohen's Kappa annotator 1-2 & 63.5\%\\
    \hline
    Cohen's Kappa annotator 1-3 & 50.6\%\\
    \hline
    Cohen's Kappa annotator 2-3 & 49.2\%\\
    \hline
    Fleiss' Kappa & 54.3\%\\
    \hline
    \end{tabular}
    \label{tab:annotation_metrics}
\end{table}

\subsection{Data Labeling}
\begin{figure*}[htbp]
\centering
    
    \begin{subfigure}[m]{0.25\textwidth}
        \centering
        \includegraphics[width=\textwidth]{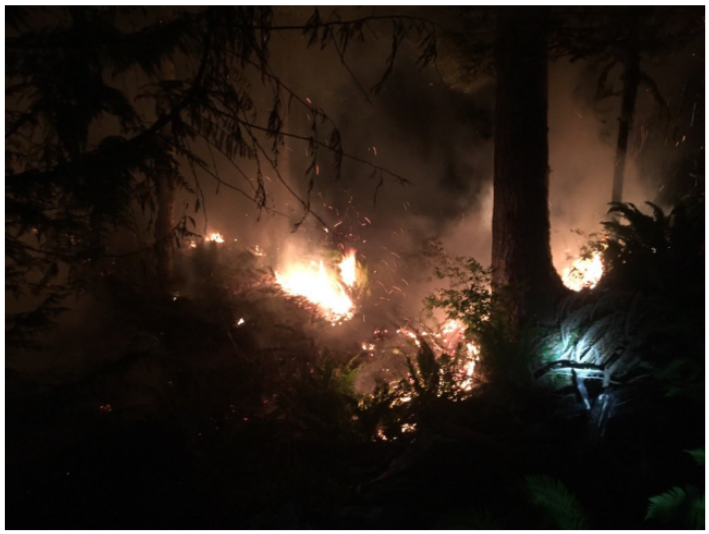}
        \caption*{\textbf{Label:} \textit{Reports of Actions of Responders}}
        \caption{\textbf{Text:} PAFD is on the scene of a fire in Roger Creek Park. There is no threat to structures at this time. \#portalberni \#BCWildfire}
        \label{fig:2-sub1}
    \end{subfigure}
    \hfill
    \begin{subfigure}[m]{0.25\textwidth}
        \centering
        \includegraphics[width=\textwidth]{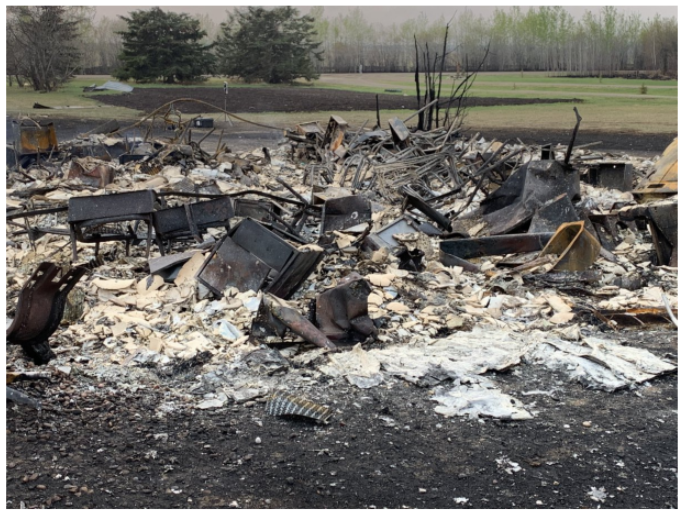}
        \caption*{\textbf{Label:} \textit{Infrastructure}}
        \caption{\textbf{Text:} The wildfires in \#AB have burned down this schoolhouse. \#ABWildfires}
        \label{fig:2-sub2}
    \end{subfigure}
    \hfill
    \begin{subfigure}[m]{0.25\textwidth}
        \centering
        \includegraphics[width=\textwidth]{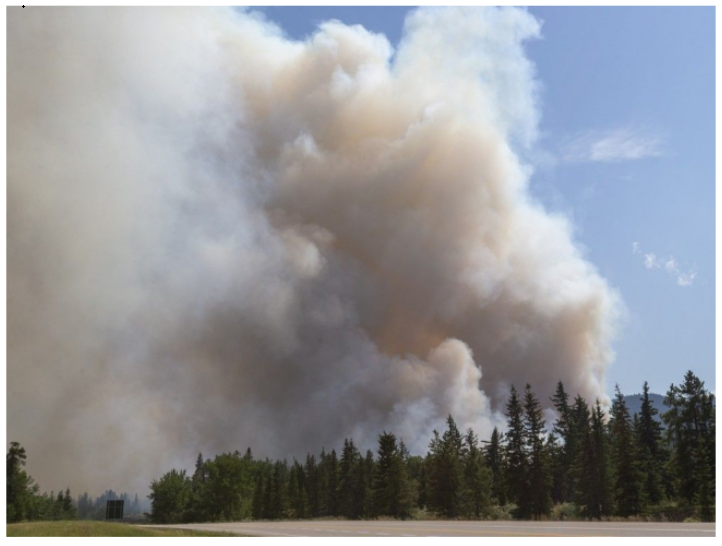}
        \caption*{\textbf{Label:} \textit{Warnings \& Status Updates}}
        \caption{\textbf{Text:} Wildfire updates, July 24: Fire reaches Jasper townsite | 'Significant loss has occured': Parks Canada | Firefighters working to protect critical infrastructure \#abfire}
        \label{fig:2-sub3}
    \end{subfigure}
    \caption{Examples of labeled posts from WildFireCan-MMD.}
    \label{fig:main-2}
\end{figure*}

Annotation was performed in Label Studio\footnote{\url{https://labelstud.io/}}, using the defined taxonomy presented in Table~\ref{tab:cat_explained}. Examples of labeled tweets are presented in Figure \ref{fig:main-2}. Each image-text pair was given a single label. In cases where the label could differ depending on whether text or image was being considered, preference was given to the modality that provided greater value for that particular sample. For example, in Figure \ref{fig:2-sub3} the image might be labeled ‘Smoke \& Air Quality’, but the text might be ‘Warnings \& Status Updates’. In this case, ‘Warnings \& Status Updates’ would be chosen as the label as the text contains more valuable information than the image alone. If neither contained detailed information about the time or location like the text does in this example, then the sample would be labeled as ‘Smoke \& Air Quality’ as that is the only value that would otherwise be provided. The size of labeled classes is presented in Table \ref{tab:cat_explained}.
 
Samples were annotated by three independent annotators, all authors of this paper. The annotators are male, 20-30 year old, fluent in English and reside in Canada. First, initial rounds of manual labeling on small subsets of the data gave a good understanding of what was being discussed and shown in the data, and confidence was gained in the decided classes. Over the course of labeling, samples were flagged that contained personal information. A vote between the labels from the three annotators decided the final labels. Annotation agreement scores can be seen in Table \ref{tab:annotation_metrics}. A majority agreement was reached on 88.4\% of the labels. Taking a vote between all annotators for the final label and comparing the result with each annotator, we can see which annotator was closest to the voted labels. From the results, annotators 1 \& 2 had the best agreement scores. This is further demonstrated by Cohen's Kappa metric between the different pairs of annotators. Fleiss' Kappa metric however, indicates that a moderate level of agreement is achieved between the independent annotators. In cases of disagreement, the final label was chosen from the `expert' annotator (Annotator 1), who had the most experience with the dataset collection and devising the taxonomy, and also had the highest agreement score with other annotators. Additional statistics can be found in Appendix \ref{appendix:annotation}.

\section{Classification Methodology}
\label{sec:classifier-models}
This section describes a range of classifiers evaluated for the 13-class (12 categories + `Other') classification of WildFireCan-MMD. We investigate three categories of classification methods. First, classical methods such as Decision Trees (DT), Gaussian Naive Bayes (GNB), K-Nearest Neighbours (KNN) and Support Vector Machines (SVM) are used to investigate a case where labeled data is available but computation is limited. Then, several VLMs such as LLaVA, Qwen, smolVLM, and GPT-4o-mini are employed as zero-shot classifiers. This is the right choice when no training data is available.  Finally, we experimented with a fine-tuned transformer-based classifier to assess its performance in the presence of labeled data and moderate computational resources. Further information on classical methods and VLMs can be found in Appendices \ref{appendix:classical_methods} and \ref{appendix:vlms}. 

\vspace{5pt}
\noindent \textbf{Architecture of custom classifier: } 
To determine which components to use in the selected architecture, we first experimented with various unimodal classifiers for each modality, as seen in Tables \ref{tab:textonly} \& \ref{tab:imageonly}. These included CNNs and transformers for the image modality, as well as various BERT-based transformers for the text modality. After individually determining the best model for each modality (ViT for image with f1: 64.19\%, and RoBERTA for text with f1: 77.98\%), various fusion techniques found in the literature and included in Table \ref{tab:prev_auth_metrics} were tested to improve the classification. The results of these multimodal experiments can be seen in Table \ref{tab:multimodal}. More information about these experiments, as well as an ablation study, can be found in Appendix \ref{appendix:ablation}.

The best performing classifier consisted of two parallel encoders, an image and text encoder. The image encoder is ViT\footnote{'google/vit-base-patch16-384'}, and the text encoder is RoBERTa\footnote{'roberta-base'}. The output CLS token of the last hidden state from each are sent through separate projection layers to reduce the dimensionality from 768 to 512. The output is passed to the transformer fusion module consisting of two encoder layers, each with d\_model=512, nhead=8, dim\_feedforward=2048, dropout=0.2. A final classification head projects the dimension from 512 to 256, which is followed by ReLU activation and dropout=0.2, then finally from 256 to num\_classes. Figure \ref{fig:model} shows the structure of this architecture.

\vspace{5pt}
\noindent \textbf{Training and Evaluation:} Three data splits with stratified sampling were used to train the model and report the results. The random seeds from splitting the data were kept for reproducibility. The hyperparameters used were a batch size of 8 and an image size of ($384\times{384}\times{3}$). Text sequences are padded to a max length of 144. Adam optimizer is used with a learning rate of $1e^{-5}$, weight\_decay=0.01, and cross-entropy loss function. The model is fully trained after 10 epochs. The best model is saved based on f-weighted-average. The model was tested on the held-out test set (938 samples).

\begin{table}[ht] 
\centering
\scriptsize
\caption{Text-only classification results.}
\begin{tabular}{|c|c|c|}
\hline
\textbf{Text Model} & \textbf{Notes} & \textbf{F1-Weighted} \\
\hline
'bert-base-cased' & mean-pooling & 0.7779 \\
'bert-base-cased' & CLS & 0.7793 \\
"bertweet-base" & mean-pooling & 0.7900 \\
"bertweet-base" & CLS & 0.7927 \\
"ModernBERT-base" & mean-pooling & 0.7934 \\
"ModernBERT-base" & CLS & 0.7862 \\
"roberta-base" & robertaforseq & 0.7718 \\
"roberta-base" & mean-pooling & 0.7934 \\
"roberta-base" & mean-pooling, half-frozen & 0.7940 \\
"roberta-base" & mean-pooling, custom vocab & 0.7209 \\
"roberta-base" & CLS & 0.7798 \\
"roberta-base" & CLS, half-frozen & \textbf{0.8008} \\
\hline
\end{tabular}
\label{tab:textonly}
\end{table}

\begin{table}[ht] 
\centering
\scriptsize
\caption{Image-only classification results.}
\begin{tabular}{|c|c|c|c|}
\hline
\textbf{Image Model} & \textbf{Notes} & \textbf{F1-Weighted} \\
\hline
DenseNet121 & ImageNet weights & 0.5733 \\
InceptionV3 & ImageNet weights & 0.5804 \\
ResNet50 & ImageNet weights & 0.5841 \\
VGG16 & ImageNet weights & 0.5993 \\
"facebook/deit-base-patch16-384" & CLS & 0.6260 \\
"google/vit-base-patch16-384" & half-frozen, CLS & 0.6290 \\
"google/vit-base-patch16-384" & mean-pooling & 0.6404 \\
"google/vit-base-patch16-384" & CLS & \textbf{0.6419} \\
\hline
\end{tabular}
\label{tab:imageonly}
\end{table}

\section{Experimental Results}
Tables \ref{tab:textonly} and \ref{tab:imageonly} show the unimodal text/image models experimented with. All experiments were conducted using the same train/test split.
 Experiments were done on different final feature representations by using either mean pooling of the last hidden state or the CLS token. Freezing the first half of the encoder layers was also tried as this has been shown to increase accuracy \cite{ingle2022}. All encoders had a linear layer appended for classification. For text, "roberta-base" performed best when using just the CLS token for classification and freezing the first half of the layers, as shown in Table \ref{tab:textonly}. Later experimentation found that the unfrozen version performed better in the final classifier. For images, "vit-base-patch16-384" performed best when using just the CLS token for classification without any freezing. Many of the fusion experiments tested, as well as some architectures from other studies can be seen in Table \ref{tab:multimodal}. The same train/test split was used as before for unimodal classification. The best result, belonging to the selected classifier, is shown in bold. Table \ref{tab:metrics_all} shows the calculated f-scores for the VLMs, selected and baseline classifiers per class and as an average over all classes. Appendix \ref{appendix:confusion} shows the confusion matrices of these classifiers. All metrics reported in these tables use 'weighted' F1-score to represent the class imbalance observed in the dataset.

\begin{table*}[h]
\centering
\scriptsize
\caption{Multimodal classification results.}
\begin{tabular}{|>{\centering\arraybackslash}p{2cm}|>{\centering\arraybackslash}p{0.6cm}|>{\centering\arraybackslash}p{1.8cm}|p{8cm}|>{\centering\arraybackslash}p{0.7cm}|}
\hline
\textbf{Text Model} & \textbf{Image Model} & \textbf{Fusion Type} & \textbf{Notes} & \textbf{F1} \\
\hline
RoBERTa & ViT & Cross Attention & encoders pretrained individually, q=image k=text v=text & 0.7855 \\
RoBERTa & ViT & Cross Attention & encoders pretrained individually, q=text k=image v=image & 0.6489 \\
RoBERTa & Coca & Concatenative & COCA/RoBERTa individually aligned w/ description of the label & 0.1587 \\
RoBERTa+BiLSTM & ViT & BiLSTM+BiLinear & Koshy's model \cite{Koshy2023} & 0.6149 \\
CLIP & CLIP &  & "openai/clip-vit-base-patch32" & 0.7072 \\
VILT & VILT &  & "dandelin/vilt-b32-finetuned-nlvr2" & 0.7638 \\
RoBERTa & ViT & Concatenative & ViT/RoBERTa individually aligned w/ description of the label & 0.6507 \\
RoBERTa & ViT & Transformer & transformer fusion: h\_dim 1024 and 0.1 dropout & 0.8263 \\
RoBERTa & ViT & Transformer & transformer fusion: h\_dim 2048 and 0.2 dropout & \textbf{0.8364} \\
RoBERTa & ViT & Transformer & RoBERTa/ViT fully frozen & 0.6164 \\
RoBERTa & ViT & Concatenative & stopwords removed from text before RoBERTa & 0.7870 \\
RoBERTa & ViT & Concatenative & CLS into single classification layer & 0.8249 \\
RoBERTa x2 & ViT & Concatenative & extract text from images with OCR & 0.8102 \\
ModernBERT & ViT & Transformer & transformer fusion: hidden dim 2048 and 0.2 dropout & 0.8149 \\
ModernBERT & ViT & Concatenative & CLS into single classification layer & 0.7966 \\
RoBERTa & ViT & Concatenative & mean-pooling of last hidden states into linear classification layer & 0.7911 \\
RoBERTa & ViT & Concatenative & roberta:mean, vit:cls & 0.8028 \\
RoBERTa & ViT & Transformer & transformer fusion: h\_dim 2048 and 0.2 dropout RoBERTa:mean, ViT:cls & 0.8223 \\
RoBERTa & ViT & Cross Attention & same as proposed but with cross attention instead of transformer & 0.8208\\
\hline
\end{tabular}
\label{tab:multimodal}
\end{table*}

\begin{table*}[h]
\centering
\tiny
\renewcommand{\arraystretch}{1.5} 
\setlength{\tabcolsep}{5pt} 
\caption{F1-scores per-class for zero-shot VLMs, baseline and selected classifiers. 
}
\label{tab:metrics_all}
\begin{tabular}{l|cccc|c|cccc}
\hline
\rowcolor[gray]{0.8} 
\textbf{} & \multicolumn{4}{c|}{\textbf{Baseline Classifiers}} & \multicolumn{1}{c|}{\textbf{Classifier}} & \multicolumn{4}{c}{\textbf{zero-shot VLMs}} \\
\rowcolor[gray]{0.8} 
\textbf{Class} & \textbf{DT} & \textbf{GNB} & \textbf{KNN} & \textbf{SVM w/ PCA} &  & \textbf{smolVLM} & \textbf{LLaVA} & \textbf{Qwen} & \textbf{GPT-4o-mini} \\
\hline
\textbf{Evacuees}                         & 39.34±7.08 & 45.34±2.26 & 45.78±4.64 & 31.65±4.69 & 88.55±3.20 & 25.84 & 17.65 & 12.70 & 37.21 \\
\rowcolor[gray]{0.95} 
\textbf{General Information}              & 27.04±6.43 & 39.11±3.73 & 36.01±0.56 & 33.64±6.32 & 82.04±3.54 & 00.00 & 19.05 & 21.88 & 20.37 \\
\textbf{Preparedness}                     & 50.27±4.05 & 65.04±4.15 & 56.25±9.15 & 52.24±10.58 & 66.02±4.10 & 22.58 & 57.14 & 67.69 & 62.50 \\
\rowcolor[gray]{0.95} 
\textbf{Weather Reports}                  & 51.48±6.89 & 61.22±3.86 & 63.70±3.44 & 65.37±2.89 & 89.01±4.51 & 00.00 & 64.52 & 50.00 & 62.39 \\
\textbf{Warnings \& Status Updates}       & 43.70±3.16 & 46.43±3.48 & 58.67±0.92 & 56.20±1.21 & 93.19±2.02 & 00.00 & 34.94 & 42.73 & 48.00 \\
\rowcolor[gray]{0.95}  
\textbf{Reports of Actions of Responders} & 47.50±4.05 & 51.49±2.81 & 54.84±5.75 & 64.06±2.15 & 81.10±2.14 & 00.00 & 26.09 & 63.89 & 69.09 \\
\textbf{Infrastructure}                   & 55.13±8.21 & 67.51±6.28 & 69.16±6.58 & 66.01±8.15 & 87.88±1.86 & 00.00 & 36.36 & 50.00 & 37.50 \\
\rowcolor[gray]{0.95} 
\textbf{Political}                        & 50.04±4.69 & 69.18±1.74 & 54.64±1.51 & 58.89±3.39 & 83.73±0.29 & 00.00 & 69.09 & 85.25 & 77.86 \\
\textbf{Insurance}                        & 86.66±1.52 & 91.25±1.31 & 90.57±3.62 & 92.09±1.12 & 87.01±1.22 & 93.75 & 88.52 & 90.91 & 86.15 \\
\rowcolor[gray]{0.95} 
\textbf{Advertisement}                    & 48.80±6.41 & 49.56±7.07 & 65.92±6.43 & 66.11±3.23 & 89.37±1.71 & 00.00 & 08.33 & 45.16 & 73.17 \\
\textbf{Smoke and Air Quality}            & 58.47±2.82 & 72.05±3.92 & 75.19±1.94 & 76.84±1.21 & 84.20±6.11 & 45.48 & 72.46 & 65.96 & 75.38 \\
\rowcolor[gray]{0.95} 
\textbf{Support}                          & 47.99±10.26 & 60.76±4.66 & 47.53±6.10 & 43.86±6.57 & 77.72±2.26 & 05.86 & 27.27 & 43.84 & 56.52 \\
\textbf{Other}                            & 46.37±2.53 & 56.17±1.76 & 48.85±2.65 & 58.93±1.65 & 81.29±1.00 & 25.36 & 05.22 & 52.46 & 57.58 \\
\rowcolor[gray]{0.8} 
\hline
\textbf{F1 Weighted Average}              & 50.88±1.11 & 60.77±1.45 & 61.22±1.23 & 62.21±0.44 & 84.48±0.69 & 19.86 & 44.69 & 55.94 & 61.17 \\
\hline
\end{tabular}
\end{table*}

The custom classifier achieved the highest weighted average F1-score (84.48\%), outperforming both zero-shot vision-language models and traditional baseline classifiers across all classes. From the results in Table~\ref{tab:metrics_all} and Appendix~\ref{appendix:confusion}, the \textit{‘Preparedness’} class had the lowest F1-score under the custom classifier (66.02±4.10\%), likely due to high variability in both imagery and associated text. On the other hand, the \textit{‘Warnings \& Status Updates’} class achieved the highest performance (93.19±2.02\%), reflecting the strong visual and textual consistency across samples. Other high-performing classes include \textit{‘Advertisement’} (89.37±1.71) \textit{‘Weather Reports’} (89.01±4.51\%), and \textit{‘Evacuees’} (88.55±3.20\%), which may be more visually or textually distinct. In contrast, mid-range scores are seen in classes like \textit{‘Support’} (77.72±2.26\%) and \textit{‘Political’} (83.73±0.29\%), which often contain dense textual overlays that may not align well with the image modality.

\section{Discussion}

\vspace{5pt}
\noindent \textbf{Task-specific fine-tuning is critical: }Table~\ref{tab:metrics_all} shows that classical models fine-tuned on labeled data generally outperform zero-shot VLMs, with the custom model achieving ~23\% higher f-score than VLMs. Despite a modest dataset, results affirm the value of targeted annotation and model tuning, echoing findings in literature presented in Appendix~\ref{appendix:litreview}.

\vspace{5pt}
\noindent \textbf{VLMs involve tradeoffs: }Zero-shot VLMs, while being a practical option when no labeled data is available, pose challenges in accuracy and cost. GPT-4o-mini offers strong results, but is costly and lacks transparency. Open-source models like LLaVA, Qwen, and smolVLM are more accessible but generally underperform, though Qwen is a notable exception, and trails GPT-4 by just 5\%. Overall, VLMs may offer a solution when labeled data is unavailable, but their performance is comparable to traditional models such as SVM and KNN, which require training data but can be implemented with minimum computation.  

\vspace{5pt}
\noindent \textbf{Future Work: }The dataset exhibits a significant class imbalance, and future work may involve steps to address this issue. Utilizing the unlabeled data paired with a labelled dataset in a semi-supervised setup is a promising avenue for future research \cite{sirbu2022, sirbu2025}. Fine-tuning of VLMs is also a promising avenue which should be explored in future works \cite{duan2024cityllava, zhai2024fine}.

\section{Analysis on Trends in Data}
\label{sec:trend}
This section describes the continuation of the analysis from the work by \citet{sherritt2025cai}. Here, we present an analysis of an additional 46,279 posts from X, posted between 2018 and 2024, using the custom classifier described in Section \ref{sec:classifier-models}. The collection methodology was the same as explained in section \ref{subsec:collection}, where tweets were gathered using a curated set of wildfire-related hashtags specific to each year, selected based on known wildfire events and hotspot locations across affected regions. This was combined with a keyword-based search, and then the data was combined, and duplicates were removed. The full set of search queries can be seen in Appendix \ref{sec:queries}. 

\begin{figure}
\centering
    \includegraphics[width=0.9\linewidth]{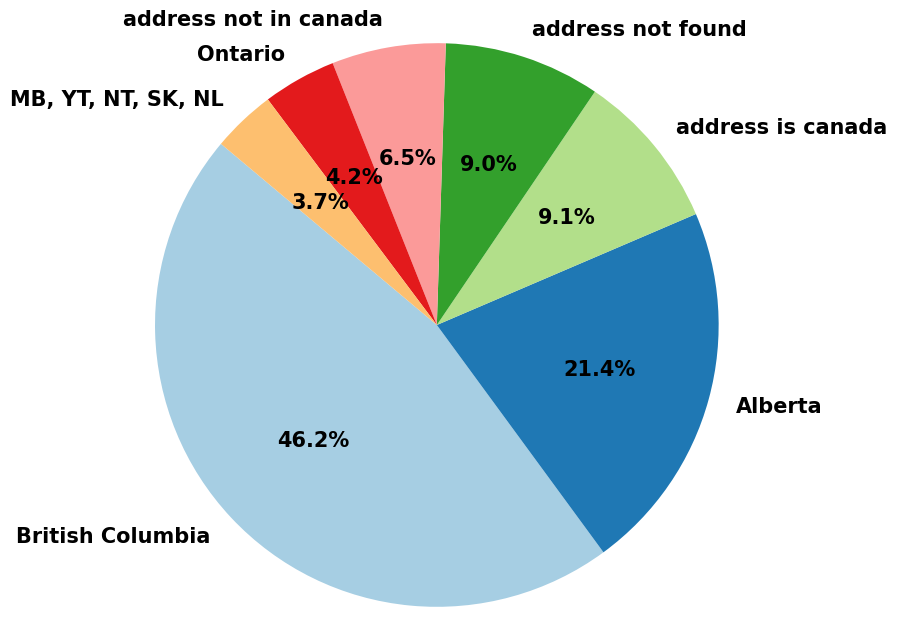}
    \caption{Distribution of collected tweets based on author location.}
    \label{fig:region}
\end{figure}

\begin{figure*}[ht]
  \centering
  \begin{subfigure}{0.44\textwidth}
    \centering
    \includegraphics[width=\linewidth]{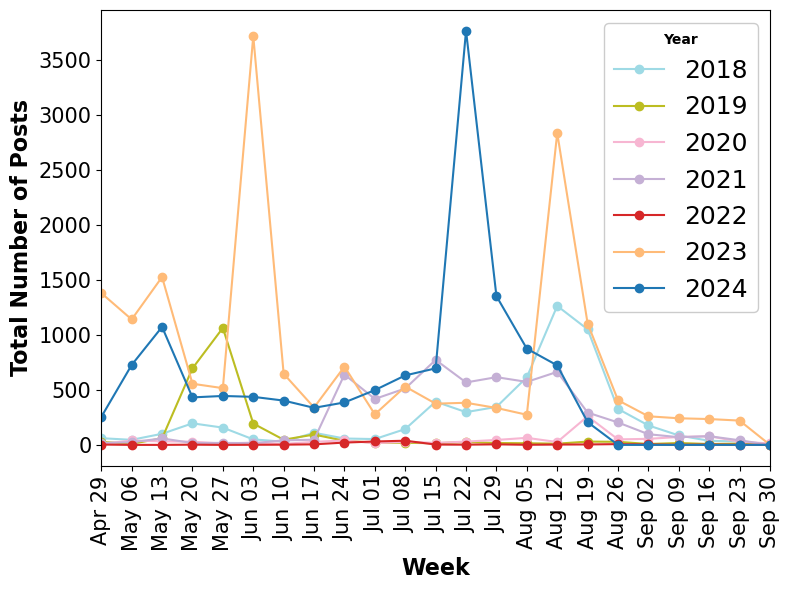}
    \label{fig:postcounts}
  \end{subfigure}\hfill
  \begin{subfigure}{0.55\textwidth}
    \centering
    \includegraphics[width=\linewidth]{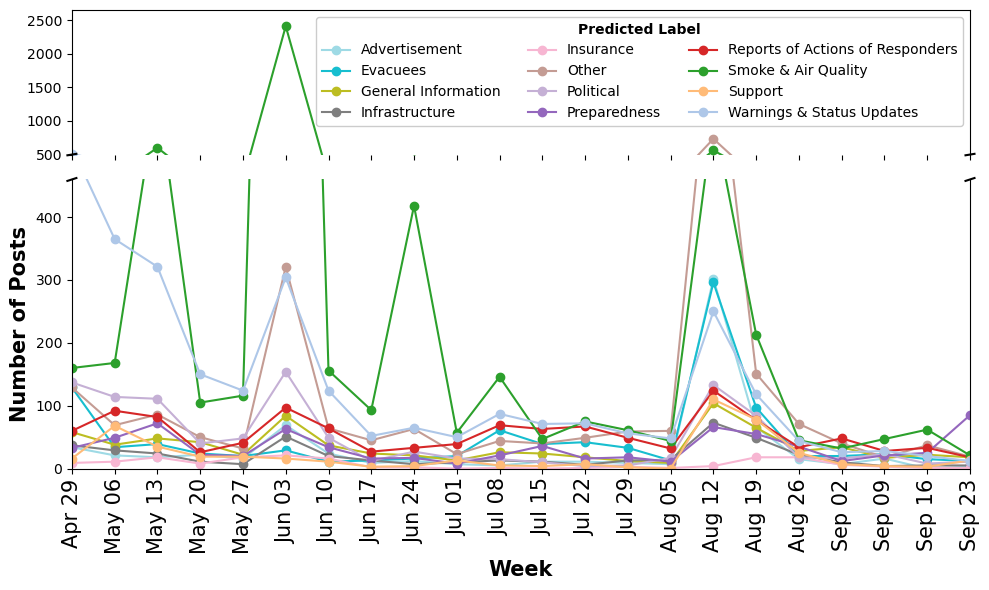}
    \label{fig:2023countsbyclass}
  \end{subfigure}
  \caption{Weekly post distributions: (left) overall yearly trend, (right) class-wise distribution in 2023.}
  \label{fig:weekly_post_distributions}
\end{figure*}

The author locations were extracted from the samples with the goal of obtaining the province-level region from which each sample was posted. These could be either automatically generated or user-provided. The queries were run through OSMNX\footnote{\url{https://osmnx.readthedocs.io/en/stable/}}, which allowed for the automated search of cities or towns that users provided. If a user's location was not found, or not in Canada, it was labeled as such. The distribution of posts by province is shown in Figure \ref{fig:region}, indicating that the majority of the data originates from British Columbia and Alberta. Additional information about this data, including a word cloud of the text, the frequency distribution of the posts by word count, can be found in Appendix \ref{appendix:analysis}. 

Using the custom classifier, predictions were obtained for the new data. This automated classification approach allowed for efficiency in processing the large volume of newly collected data without having to manually label each individual sample. This of course, will not uphold the same gold standard of labeling that a human annotator could achieve. However, while individual classifications may not always be accurate, general trends in the data can be analyzed and linked to real events.

Figure \ref{fig:weekly_post_distributions} (left) shows the total number of posts per week sorted by year. This demonstrates a significant event occurred the week of June 3, 2023, and July 22, 2024, due to several major fires in the country \cite{reuters2023bcbfires, cbc2024jasperfire}. Figure \ref{fig:weekly_post_distributions} (right) illustrates weekly trends in predicted labels during 2023. Sharp spikes in \textit{Smoke and Air Quality} and \textit{Warnings \& Status Updates} likely correspond to major wildfire events occurring in Alberta and BC. \textit{Warnings \& Status Updates} and \textit{Evacuees} follow similar patterns around the week of Aug 12, which corresponds to the state of emergency declared as the city of West Kelowna was evacuated. Overall, at least 35,000 people were under evacuation order and another 30,000 under evacuation alert, as of August 19 \cite{williams2023bcwildfires}. \textit{Reports of Actions of Responders} and \textit{Support} also rise during these peaks, reflecting coordinated emergency efforts.

\section{Conclusion}
Despite the existence of other natural disaster datasets from social media, these datasets often lack wildfire data from a significant period of time, typically spanning only a single month surrounding a particular event. Furthermore, even less of this data comes from Canada and is mainly restricted to the US context. To address this gap, we present WildFireCan-MMD, a wildfire-specific, multimodal dataset of social media posts curated from X, accompanied by a domain-focused taxonomy derived through multimodal topic modeling and manual annotation by multiple independent annotators. Unlike existing disaster datasets, WildFireCan-MMD is specifically designed for the Canadian wildfire context and provides a rich resource for training and evaluating specialized models. It also serves as a true test set for evaluating VLMs in a zero-shot setting. We provide the results of classifiers trained on this dataset from classical approaches to custom deep learning architectures and several VLMs, to showcase various scenarios in terms of data and computational resource availability. Our best results are obtained with an architecture that utilizes pre-trained encoders and transformer fusion for classification on this dataset, which highlights the importance of a labelled dataset. As climate-induced wildfires grow in frequency and complexity, this dataset and model support the development of robust tools for detection, coordination, and recovery. 

\section*{Acknowledgment}
This work was funded by the National Program Office of the National Research Council Canada, under the New Beginning Initiative Program.

\section*{Limitations}

While this study introduces a novel wildfire-specific multimodal dataset and demonstrates the effectiveness of fine-tuned task-specific models, several limitations should be acknowledged.

The first limitation is the scope and size of the dataset. Although WildFireCan-MMD provides a targeted and realistic benchmark for wildfire-related content, its size remains modest compared to large-scale datasets used to train or evaluate deep learning models. This limits the diversity of linguistic and visual expressions captured and may affect the generalizability of these results to other wildfire events or regions.

Platform and geographic bias might be present in our data. The dataset is sourced exclusively from the X platform and focuses on Canadian wildfires. This introduces both platform-specific and regional biases in language use, imagery, and public discourse. As a result, models trained on this dataset may not perform as well when applied to other social platforms or international wildfire contexts.

Moreover, zero-shot evaluation of VLMs introduces limitations. Prompt engineering for zero-shot setups is a heuristic process, and better prompts might yield different results.

Finally, there is a broader risk of misapplication. Deploying these models in real-world disaster contexts without appropriate oversight, interpretability mechanisms, or human-in-the-loop verification may lead to misclassification, misinformation amplification, or missed signals, potentially affecting emergency coordination and public trust.

 \bibliography{custom}

\appendix

\section{Annotation}
\label{appendix:annotation}
Figure \ref{fig:annotation_distrib} shows the distribution of the labels over the classes by each independent annotator.
\setcounter{table}{0}
\counterwithin{table}{section}
\renewcommand\thetable{\thesection.\arabic{table}}
\setcounter{figure}{0}
\counterwithin{figure}{section}
\renewcommand\thefigure{\thesection.\arabic{figure}}
\begin{figure}[h]
  \centering
  \includegraphics[width=\linewidth]{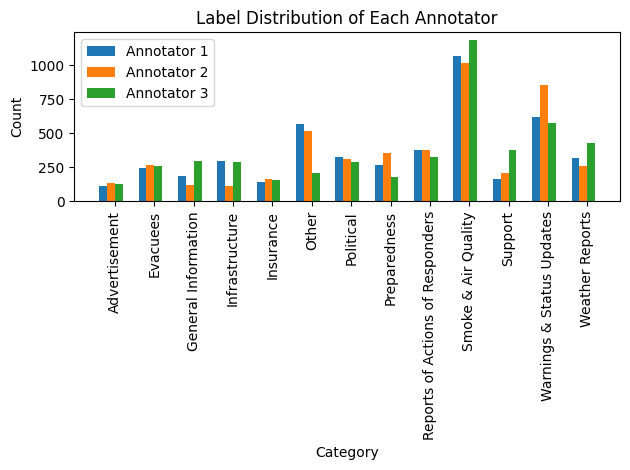}
  \caption{Counts of labels by each annotator}
  \label{fig:annotation_distrib}
\end{figure}

\section{Topic Modeling}
\label{appendix:bertopic}

Figure \ref{fig:main-1} shows examples of topics extracted by the unsupervised BERTopic. We present representative images and corresponding keywords generated by BERTopic for each example. As seen in the captions of the image, some of the topics are not useful and present non-coherent concepts, while others are useful and relevant to wildfires. 

\setcounter{table}{0}
\counterwithin{table}{section}
\renewcommand\thetable{\thesection.\arabic{table}}
\setcounter{figure}{0}
\counterwithin{figure}{section}
\renewcommand\thefigure{\thesection.\arabic{figure}}

\begin{figure*}[htbp]
    \centering
    \begin{subfigure}[m]{0.3\textwidth}
        \centering
        \includegraphics[width=\textwidth]{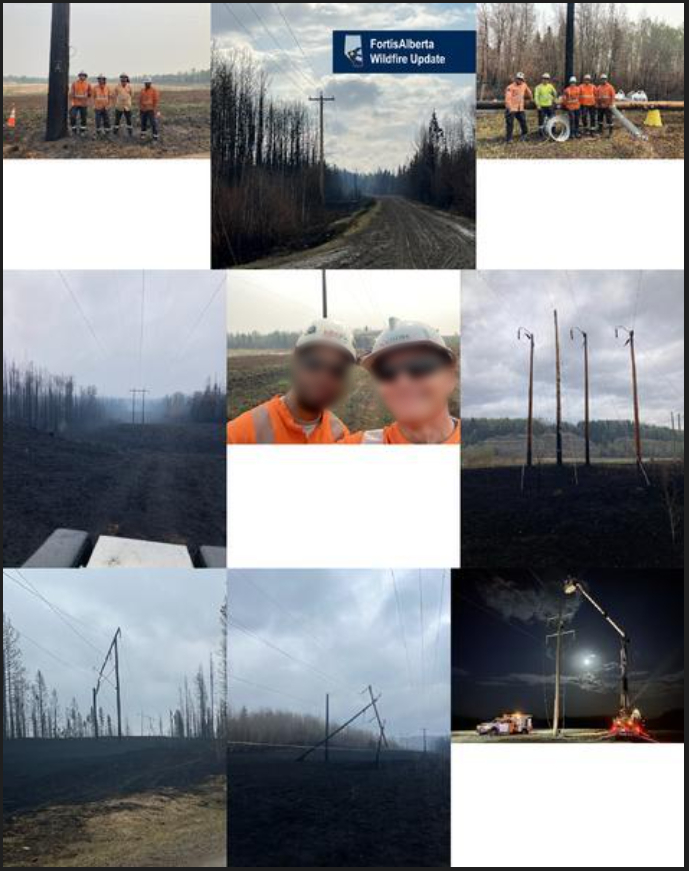}
        \caption*{\textbf{Keywords:} [crews, safe, poles, damaged, continue, lines, working restore, power, conditions, restore]}
        \caption{\textbf{Useful}}
        \label{fig:sub1}
    \end{subfigure}
    \hfill 
    \begin{subfigure}[m]{0.25\textwidth}
        \centering
        \includegraphics[width=\textwidth]{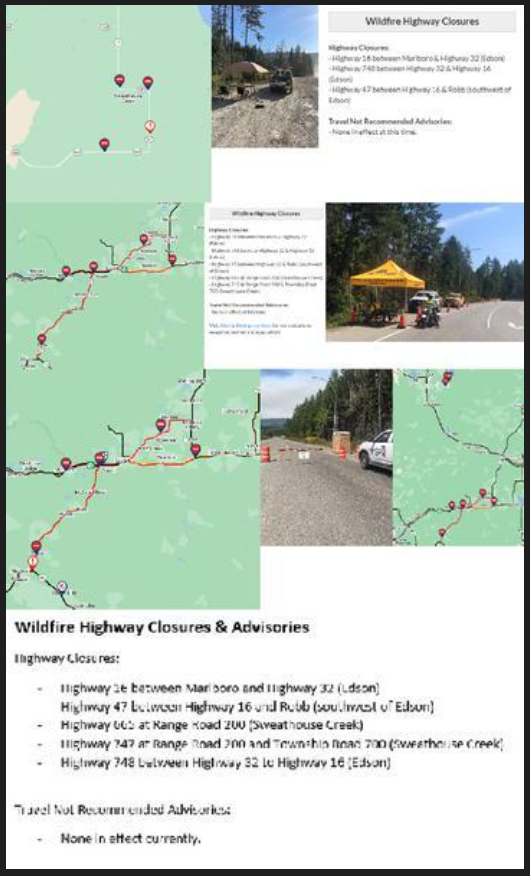}
        \caption*{\textbf{Keywords:} [highway, closed, closures, abroads, detour, road, route, bchwy4, cameron, cameron lake]}
        \caption{\textbf{Useful}}
        \label{fig:sub2}
    \end{subfigure}
    \hfill
    \begin{subfigure}[m]{0.35\textwidth}
        \centering
        \includegraphics[width=\textwidth]{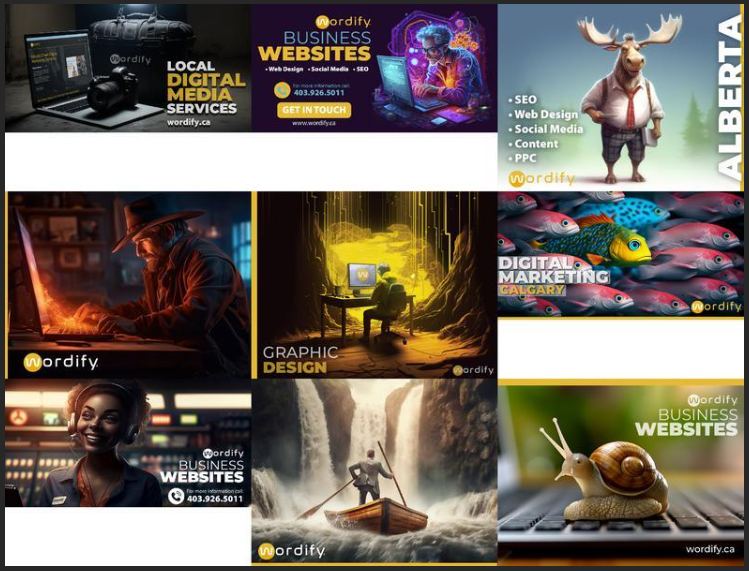}
        \caption*{\textbf{Keywords:} ['yycnow', 'yycbusiness', 'yycbusiness yycnow', 'yyc yycbusiness', 'abfires abwildfire', 'calgary', 'startups', 'wordify', 'smallbusiness', 'smallbusiness startups']}
        \caption{\textbf{Non-useful}}
        \label{fig:sub3}
    \end{subfigure}
    \caption{Examples of useful/non-useful topics from topic modeling on our dataset.}
    \label{fig:main-1}
\end{figure*}

\section{Prompts for Zero-Shot Classification}
\label{appendix:prompts}
The full prompt used to evaluate the aforementioned VLMs is shown in Table \ref{tab:sys_usr_prompt}.
\setcounter{table}{0}
\counterwithin{table}{section}
\renewcommand\thetable{\thesection.\arabic{table}}
\setcounter{figure}{0}
\counterwithin{figure}{section}
\renewcommand\thefigure{\thesection.\arabic{figure}}
\begin{table}[h]
\footnotesize
  \caption{System and User prompt used for VLMs.}
  \label{tab:sys_usr_prompt}
  \begin{tabular}{p{0.9\linewidth}}
    \toprule
    \begin{minipage}[t]{\linewidth}
    \textit{\textbf{System Prompt}: \\
    "You are an assistant who is being given an image and text pair as a Twitter post, which was created during a natural disaster event. Your task is to use information from both the text and the image to decide which option the post should be labeled as. You must pay close attention to each option when deciding which label to use."}
    \end{minipage} \\
    \midrule
    \begin{minipage}[t]{\linewidth}
    \textit{\textbf{User Prompt}: \\
    "Which option should this post be labeled as? \\
    A. Evacuees (information relating to evacuees, their movements, needs, location, etc) \\
    B. General Information (GENERAL facts about the wildfire situation, hectares burned) \\
    C. Preparedness (information for the general public to prepare themselves and property for wildfires) \\
    D. Weather Reports (information relating to the weather, satellite imagery, radar imagery) \\
    E. Warnings \& Status Updates (warnings/updates to the public from authoritative bodies, fire bans, specific information relating to a certain time or area) \\
    F. Reports of Actions of Responders (prescribed burns, responders responding to a specific location) \\
    G. Infrastructure (road closures, damaged buildings or property, traffic) \\
    H. Political (mentions of political or public figures or parties) \\
    I. Insurance (mentions of insurance) \\
    J. Advertisement (information about restaurants, food, apps or services) \\
    K. Smoke \& Air Quality (information about smoke or air quality, masks, breathing) \\
    L. Support (information about financial, mental health, or other types of support for people) \\
    M. Other (the post does not fit well in one of the previous categories) \\
    You may only answer with the chosen option's letter."}
    \end{minipage} \\
    \bottomrule
  \end{tabular}
\end{table}

\section{Baseline Classifiers}
\label{appendix:classical_methods}
To establish a baseline, several classical methods were evaluated on the dataset, namely DT, GNB, KNN, SVM. For encoding data into numerical features for these methods, (pretrained, but not finetuned) RoBERTa and ViT were used to extract feature sets for each modality. The concatenated feature sets from the last hidden state were saved. Principle Component Analysis (PCA) was used on the feature sets to reduce the dimensionality from 768 to 250 for the SVM method as it improved the results. For DT, class\_weight='balanced' was used. For KNN, n\_neighbors=1 was used. All other hyperparameters remained default from Scikit-learn\footnote{https://scikit-learn.org/stable/}. Three different random splits were used to train and test all methods to report a confidence interval.

\section{VLMs as Zero-Shot Classifiers:}
\label{appendix:vlms}
Four VLMs were considered for experimentation, GPT-4o-mini \cite{OpenAI2023}, LLaVA \cite{Liu2023}, smolVLM \cite{smolvlm2025}, Qwen2.5-VL \cite{qwen2025} and tested in a zero-shot setting. Classification prompts for all four VLMs were created. With each query to the model, a list of all categories and their descriptions are given as a prompt along with a data instance (text + image) to classify. The prompt specifies that the model's task is to choose one of the provided categories and respond only with its corresponding letter. The system and user prompts can be seen in Appendix \ref{appendix:prompts} and are the same for all VLMs. The same hyperparameters are used for all VLMs with temperature=0.1 num\_beams=1, max\_new\_tokens=1024. The same test set was used for all VLMs and consisted of a stratified split. This single split was discovered to be a good split from previous experimentation with baseline and custom classifiers, and is one of the same three splits used for these classifiers.

\vspace{5pt}
\noindent{\textbf{SmolVLM-2B:}} Being the smallest of the VLMs experimented with, with only 2.2B parameters and <10GB of GPU ram, experiments were able to be performed locally without quantization. 

\vspace{5pt}
\noindent{\textbf{Qwen2.5-VL-7B:}} Qwen is an intermediately-size VLM with around 7B parameters. Being too large for the hardware available, manual quantization using BitsandBytes was performed to shrink the precision down to 4-bit to be used locally. 

\vspace{5pt}
\noindent{\textbf{LLaVA-v1.5-13B:}} Also considered an intermediately-sized option, the 13B 4-bit quantized version of LLaVA was used due to hardware constraints. Using a combination of a publicly available Colab notebook \footnote{\url{https://github.com/camenduru/LLaVA-colab}}, the author of which performed the quantization, and the original code from the open-source LLaVA repository \footnote{\url{https://github.com/haotian-liu/LLaVA}}, the model was able to be used with Google Colab.

\vspace{5pt}
\noindent{\textbf{GPT-4o-mini:}} The largest of the VLMs tested, a paid API service had to be used for querying GPT-4o-mini \footnote{version:GPT-4o-mini-2024-07-18}, and it is also the only closed-source model tested.

\vspace{5pt}
For each method, the responses are saved as a file where each line contains the model's answer supplied as a letter from the options shown in Table \ref{tab:sys_usr_prompt}. This then allows for the calculation of metrics.

\section{Related Architectures}
\label{appendix:litreview}
Table \ref{tab:prev_auth_metrics} briefly describes the architectures proposed by various authors for the CrisisMMD dataset. The purpose of this table is to demonstrate the literature review used to select the classifier proposed in section \ref{sec:classifier-models} and give an idea of the expected accuracy of such a classifier on this task of wildfire social-media data.
\setcounter{table}{0}
\counterwithin{table}{section}
\renewcommand\thetable{\thesection.\arabic{table}}
\setcounter{figure}{0}
\counterwithin{figure}{section}
\renewcommand\thefigure{\thesection.\arabic{figure}}
\begin{table*}[h]
\centering
\scriptsize
\begin{tabular}{|c|c|c|c|c|}
\hline
\textbf{Paper} & \textbf{Method} & \textbf{Task 1} & \textbf{Task 2} & \textbf{Task 3} \\
\hline
Agar. 2020 \cite{Agarwal2020} & InceptionV3+RCNN+AttentionFusion & 99.00 & 99.00 & 97.00 \\
Abav. 2020 \cite{Abavisani2020} & DenseNet+BERT+CrossAttentionFusion & 89.35 & 91.82 & 70.41 \\
Ofli. 2020 \cite{Ofli2020} & VGG16+CNN+Concat.Fusion & 84.20 & 78.30 & - \\
Xukun 2020 \cite{xukun2020} & AlexNet+GRU+Concat.Fusion & 78.13 & - & - \\
Madi. 2021 \cite{Madichetty2021} & DenseNet+BERT+AverageFusion & 74.85 & - & - \\
Kumar 2022 \cite{kumar2022} & VGG16+LSTM+Concat.Fusion & 82.00 & - & - \\
Pran. 2022 \cite{pranesh2022} & ResNet+RoBERTa+TransformerAttentionFusion & 85.10 & 89.50 & - \\
Sirbu 2022 \cite{sirbu2022} & semi-supervised & 82.00 & 87.20 & 88.70 \\
Diva. 2022 \cite{Divakaran2022} & DenseNet+RoBERTa+Concat.Fusion & 89.26 & 85.87 & - \\
Kotha 2022 \cite{kotha2022} & RegNetY320+BERT+Concat.Fusion & 89.63 & 86.59 & - \\
Khat. 2022 \cite{Khattar2022} & VGG16+BiLSTM+CrossAttentionFusion & 84.08 & - & - \\
Madi. 2023 \cite{Madichetty2023} & VGG16+RoBERTa+MultiplicativeFusion & 90.21 & - & - \\
Basit 2023 \cite{Basit2023} & DeIT+ALBERT+Concat.Fusion & 79.00 & 80.00 & 86.00 \\
Koshy 2023 \cite{Koshy2023} & ViT+RoBERTa+BiLSTM+BilinearFusion & 97.14 & - & - \\
Gupta 2024 \cite{Gupta2024} & DenseNet+ELECTRA+CrossAttentionFusion & 91.20 & 93.40 & 72.20 \\
Shet. 2025 \cite{shetty2025} & ViT+RoBERTa+CrossAttentionFusion & 91.25 & 89.63 & 68.72 \\
Sirbu 2025 \cite{sirbu2025} & semi-supervised & 91.22 & 89.55 & - \\
Teng  2025 \cite{teng2025} & VGG16+BERT+CrossAttentionFusion & 81.00 & - & - \\ 
Dar   2025 \cite{Dar2025} & (ResNet50+BERT)+(Graph)+IDEAFusion & 98.23 & 90.13 & - \\
\hline
\end{tabular}
\caption{Literature review of deep-learning approaches used by other works on the CrisisMMD dataset.}
\label{tab:prev_auth_metrics}
\end{table*}

\section{Ablation \& Other Experiments}
\label{appendix:ablation}
\setcounter{table}{0}
\counterwithin{table}{section}
\renewcommand\thetable{\thesection.\arabic{table}}
\setcounter{figure}{0}
\counterwithin{figure}{section}
\renewcommand\thefigure{\thesection.\arabic{figure}}
Table \ref{tab:modelablation} shows the ablation study performed on the selected model. (1) is the result of the selected model, (2) is the result of the model with a simpler classification head consisting of a single linear layer, (3) is the model with the proposed classification head but without the fusion module. Two other random splits were later used to support the result described in section \ref{sec:classifier-models}.

\begin{table}[h] 
\centering
\small
\begin{tabular}{|c|c|c|}
\hline
\textbf{Ablation} & \textbf{Explanation} & \textbf{F1-Weighted} \\
\hline
1 & Proposed &  83.64 \\
2 & Linear class. head & 82.70 \\
3 & No fusion module & 81.71 \\
\hline
\end{tabular}
\caption{Ablation of model components.}
\label{tab:modelablation}
\end{table}

\section{Structure of Custom Classifier}
\label{appendix:modelvis}
\setcounter{table}{0}
\counterwithin{table}{section}
\renewcommand\thetable{\thesection.\arabic{table}}
\setcounter{figure}{0}
\counterwithin{figure}{section}
\renewcommand\thefigure{\thesection.\arabic{figure}}
Figure \ref{fig:model} shows the architecture of the selected approach. This model is described in detail in Section \ref{sec:classifier-models}.
\begin{figure*}[h]
  \centering
  \includegraphics[width=0.7\linewidth]{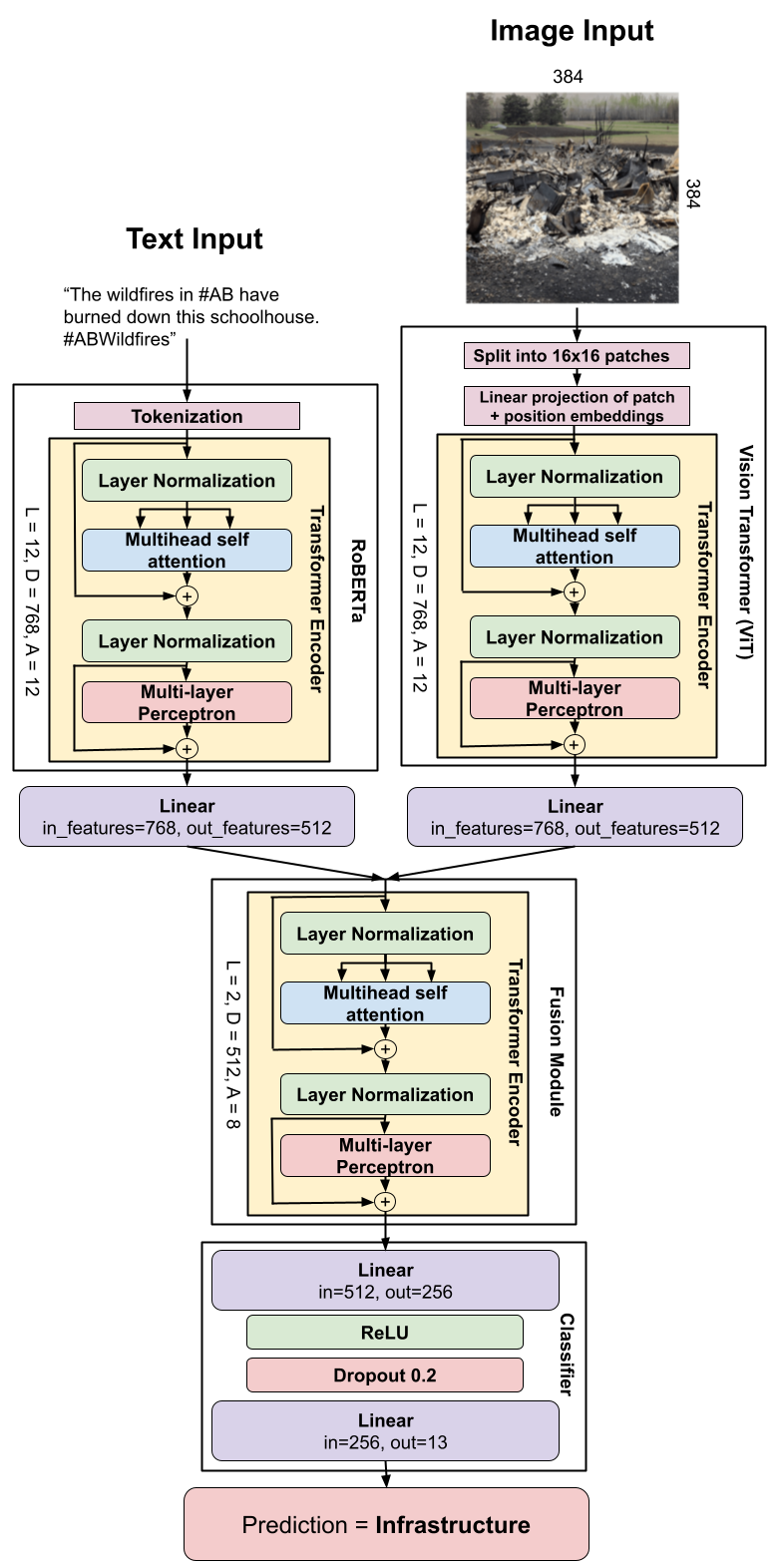}
  \caption{Selected dual-encoder architecture utilizing transformer fusion module.}
  \label{fig:model}
\end{figure*}

\section{Confusion Matrices}
\label{appendix:confusion}
\setcounter{table}{0}
\counterwithin{table}{section}
\renewcommand\thetable{\thesection.\arabic{table}}
\setcounter{figure}{0}
\counterwithin{figure}{section}
\renewcommand\thefigure{\thesection.\arabic{figure}}
For deeper insights into the classifiers performances, the confusion matrices are shown in Figures \ref{fig:classifierconfusion} and \ref{fig:vlmconfusion}.

\begin{figure*}[h]
  \centering
  \begin{subfigure}[m]{0.50\textwidth}
        \centering
        \includegraphics[trim={0cm 0cm 2cm 0cm}, clip, height=8cm]{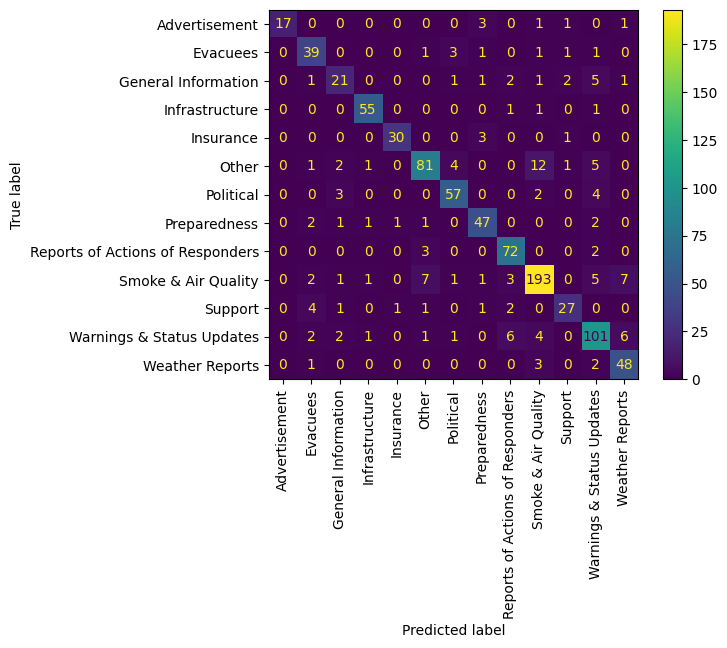} %
        \caption{Random seed: 8}
        \label{fig:classifier_8}
    \end{subfigure}
    \begin{subfigure}[m]{0.49\textwidth}
        \centering
        \includegraphics[trim={6.6cm 0cm 0cm 0cm}, clip, height=8cm]{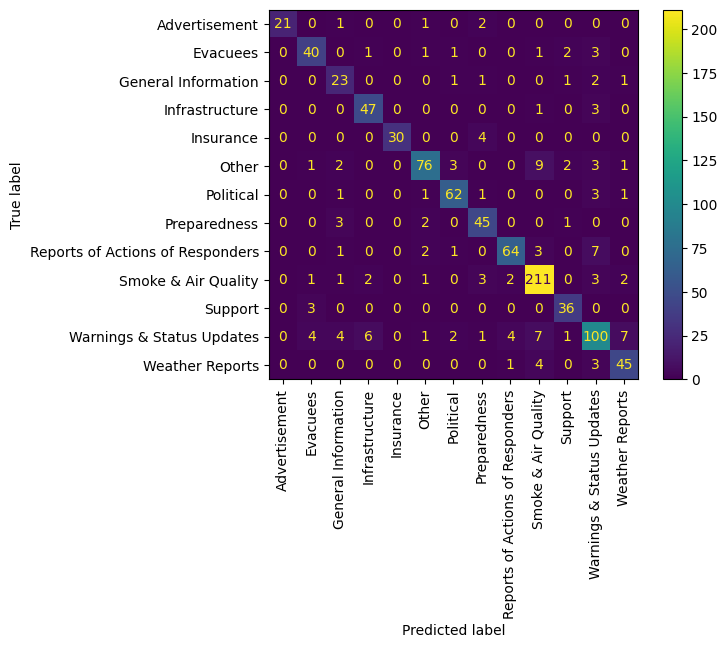}
        \caption{Random seed: 12}
        \label{fig:classifier_12}
    \end{subfigure}
    \begin{subfigure}[m]{0.49\textwidth}
        \centering
        \includegraphics[trim={0cm 0cm 0cm 0cm}, clip, height=8cm]{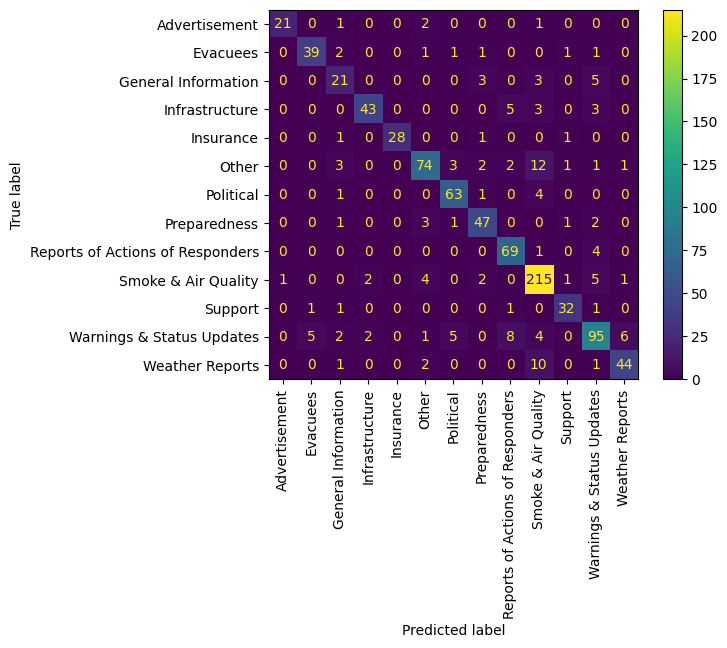}
        \caption{Random seed: 14}
        \label{fig:classifier_14}
    \end{subfigure}
    \caption{Confusion matricies for selected classifier over the different train/test splits.}
    \label{fig:classifierconfusion}
\end{figure*}

\begin{figure*}[htbp]
    \centering
    \begin{subfigure}[m]{0.50\textwidth}
        \centering
        \includegraphics[trim={0cm 9cm 0cm -1.4cm}, clip, height=6cm]{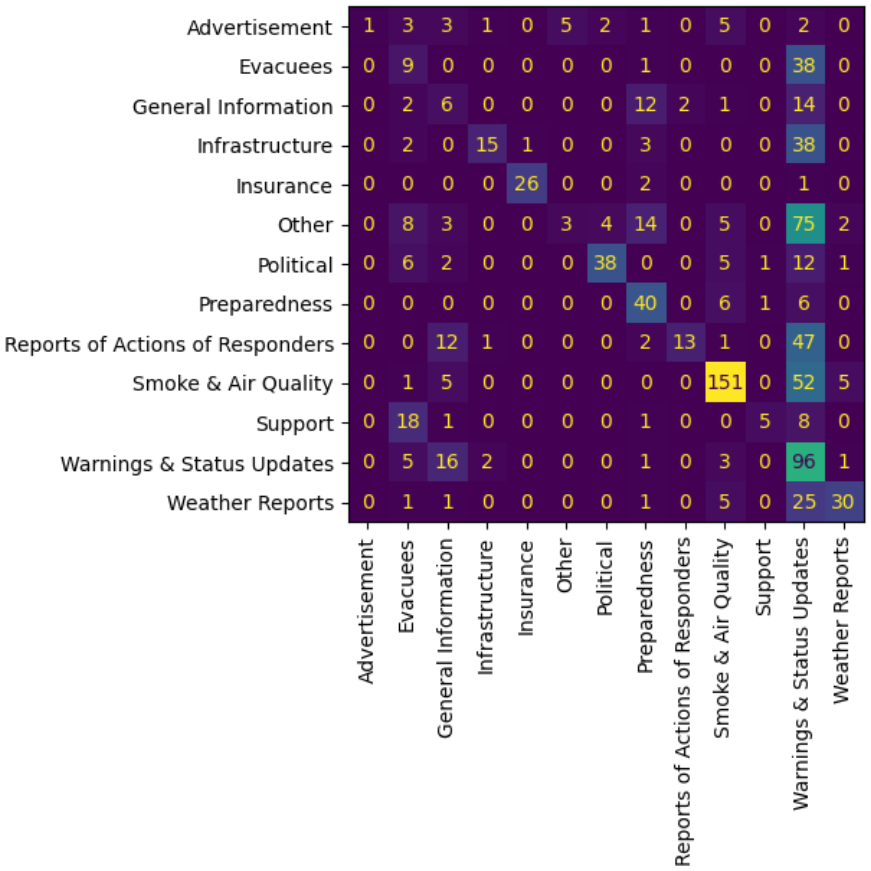} %
        \caption{LLaVA.}
        \label{fig:llavaconfusion}
    \end{subfigure}
    \begin{subfigure}[m]{0.49\textwidth}
        \centering
        \includegraphics[trim={9cm 9cm 0cm 0cm}, clip, height=5.6cm]{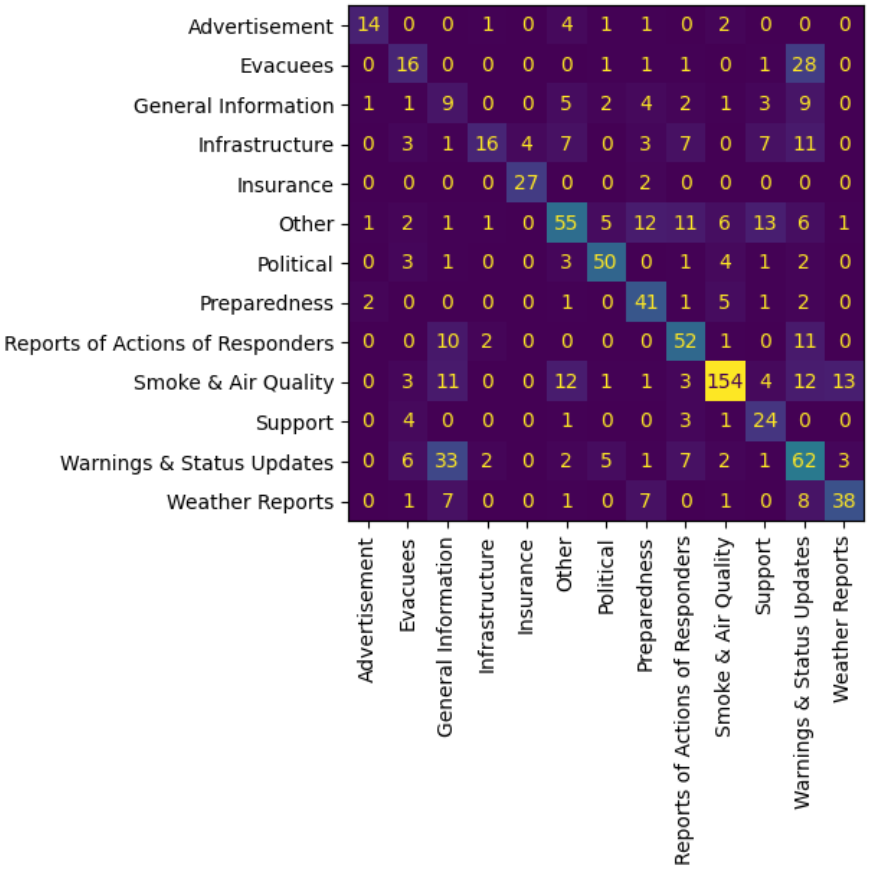}
        \caption{GPT-4.}
        \label{fig:gpt4confusion}
    \end{subfigure}
    \begin{subfigure}[m]{0.50\textwidth}
        \centering
        \includegraphics[trim={0cm 0cm 0cm 0cm}, clip, ,height=9cm]{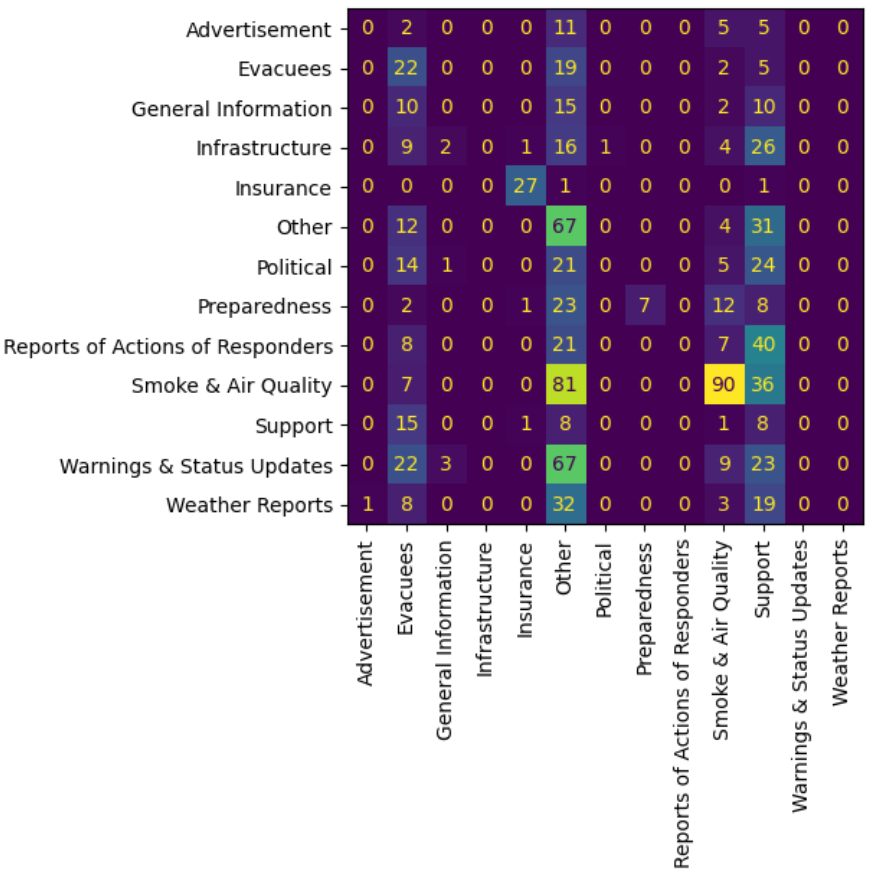}
        \caption{smolVLM.}
        \label{fig:smolvlmconfusion}
    \end{subfigure}
    \begin{subfigure}[m]{0.49\textwidth}
        \centering
        \includegraphics[trim={9cm 0cm 0cm 0cm}, clip,height=9cm]{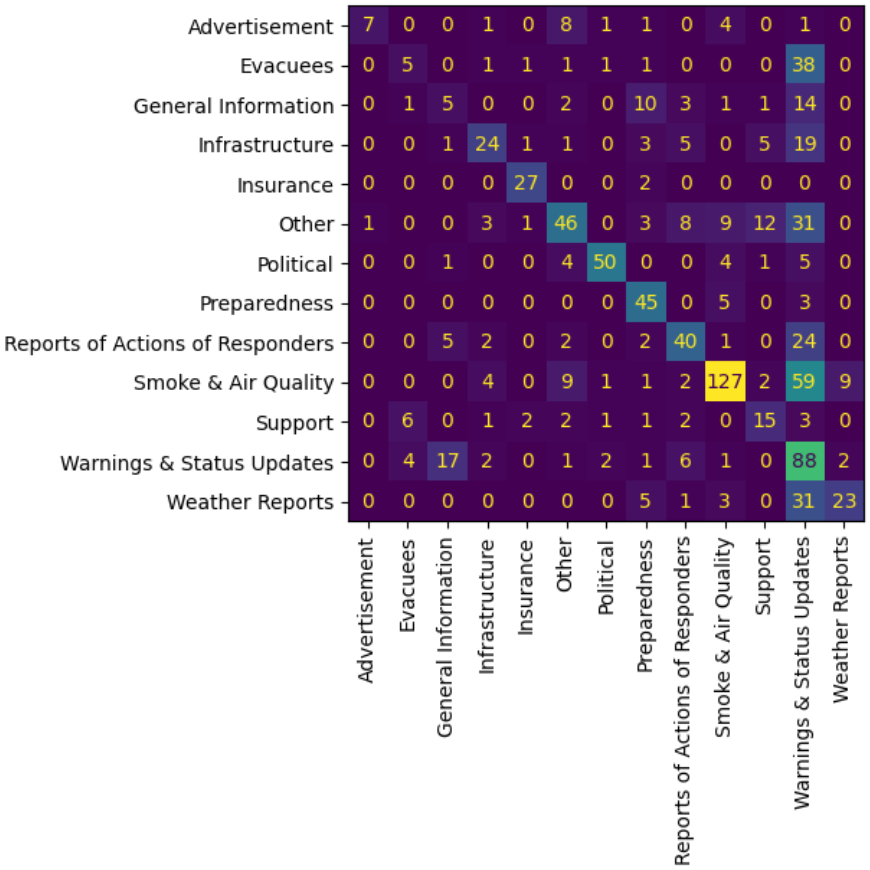}
        \caption{Qwen.}
        \label{fig:qwenconfusion}
    \end{subfigure}
    \caption{Confusion matrices of VLMs.}
    \label{fig:vlmconfusion}
\end{figure*}

\section{Queries Used for Additional Data Collection}
\label{sec:queries}
Shown here is the list of queries and keywords used to collect the additional unlabeled data discussed in Section \ref{sec:trend}.

\vspace{0.5em}
\noindent{\textbf{2018}} \\
\textit{Hashtag Query:} (\#BCwildfire OR \#britishcolumbiawildfire OR \#BCfire OR \#ABWildfire OR \#albertawildfire OR \#ABFire) \\
\textit{Keyword Query:} (alberta OR british columbia OR Prince George OR Grande Praire OR Waterton OR Bulkley Nechako OR Nadina Lake OR Kootenay OR Crowsnest Pass OR Medicine Lake OR Comstock Lake OR Tugwell Creek OR Sooke OR Nanaimo Lakes OR Tweedsmuir OR Johnny Creek OR Alkali Lake OR Lutz Creek OR Shovel Lake OR Nadina Lake OR Verdun Mountain OR Silver Lake OR Tommy Lakes OR Island Lake OR Chutanli Lake) (wildfire OR forest fire)

\vspace{0.5em}
\noindent{\textbf{2019}} \\
\textit{Hashtag Query:} (\#ABWildfire OR \#albertawildfire OR \#ABFire) \\
\textit{Keyword Query:} (alberta OR calgary OR edson OR Fort McMurray OR Grande Prairie OR High Level OR Lac La Biche OR Whitecourt OR Steen River OR Chuckegg Creek OR Peace River OR Slave Lake OR Wood Buffalo National Park) (wildfire OR forest fire)

\vspace{0.5em}
\noindent{\textbf{2020}} \\
\textit{Hashtag Query:} (\#BCwildfire OR \#britishcolumbiawildfire OR \#BCfire OR \#ABwildfire OR \#albertawildfire OR \#ABFire OR \#SKwildfire OR \#sasksatchewanwildfire OR \#SKfire OR \#YTwildfire OR \#yukonwildfire OR \#YTfire OR \#NTwildfire OR \#northwestterritorieswildfire OR \#NTfire OR \#NWTwildfire OR \#NWTfire OR \#MBwildfire OR \#manitobawildfire OR \#MBfire OR \#ONwildfire OR \#ontariowildfire OR \#QCwildfire OR \#quebecwildfire OR \#QCfire) \\
\textit{Keyword Query:} (british columbia OR alberta OR sasksatchewan OR yukon OR northwest territories OR manitoba OR ontario OR quebec) (wildfire OR forest fire)

\vspace{0.5em}
\noindent{\textbf{2021}} \\
\textit{Hashtag Query:} (\#MBwildfire OR \#manitobawildfire OR \#MBfire OR \#ONwildfire OR \#ontariowildfire OR \#SKwildfire OR \#sasksatchewanwildfire OR \#SKfire OR \#pafire OR \#ontariofire OR \#manitobafire OR \#sasksatchewanfire) \\
\textit{Keyword Query:} (manitoba OR ontario OR sasksatchewan OR british columbia) (wildfire OR forest fire)

\vspace{0.5em}
\noindent{\textbf{2022}} \\
\textit{Hashtag Query:} (\#YTwildfire OR \#yukonwildfire OR \#YTfire OR \#yukonforestfire OR \#NTwildfire OR \#northwestterritorieswildfire OR \#NTfire OR \#NWTwildfire OR \#NWTfire OR \#nwtforestfire) \\
\textit{Keyword Query:} (\#yzf OR \#nwt OR \#Yellowknife OR Yukon OR Northwest Territories OR Whitehorse OR Yellowknife OR Dawson City OR Great Slave Lake OR Norman Wells OR Inuvik OR Watson Lake OR Hay River OR Fort Smith OR Tuktoyaktuk OR Behchoko) (wildfire OR forest fire) (-alaska -Eielson -CityofNorthPole)

\vspace{0.5em}
\noindent{\textbf{2023}} \\
\textit{Hashtag Query:} (\#BCwildfire OR \#britishcolumbiawildfire OR \#BCfire OR \#ABwildfire OR \#albertawildfire OR \#ABFire OR \#SKwildfire OR \#sasksatchewanwildfire OR \#SKfire OR \#YTwildfire OR \#yukonwildfire OR \#YTfire OR \#NTwildfire OR \#northwestterritorieswildfire OR \#NTfire OR \#NWTwildfire OR \#NWTfire OR \#MBwildfire OR \#manitobawildfire OR \#MBfire OR \#ONwildfire OR \#ontariowildfire OR \#QCwildfire OR \#quebecwildfire OR \#QCfire OR \#CanadaOnFire OR \#CanadaWildfire OR \#CanadaFires OR \#CanadaIsOnFire) \\
\textit{Keyword Query:} (ontario OR quebec OR sasksatchewan OR british columbia OR manitoba OR northwest territories OR yukon OR alberta) (wildfire OR forest fire)

\vspace{0.5em}
\noindent{\textbf{2024}} \\
\textit{Hashtag Query:} (\#JasperStrong OR \#JasperWildfire OR \#JasperAB OR \#BCwildfire OR \#britishcolumbiawildfire OR \#BCfire OR \#ABwildfire OR \#albertawildfire OR \#ABFire OR \#CanadaOnFire OR \#CanadaWildfire OR \#CanadaFires OR \#CanadaIsOnFire) \\
\textit{Keyword Query:} (canada OR ontario OR quebec OR sasksatchewan OR manitoba OR northwest territories OR yukon OR british columbia OR alberta OR jasper) (wildfire OR forest fire)

\vspace{0.5em}
\noindent \textbf{Note:} Every query includes the suffix \texttt{-has:videos has:images lang:en -is:retweet -is:quote -is:reply} to filter for original English-language posts with images only.

\section{Unlabeled Dataset }
\label{appendix:analysis}
\setcounter{table}{0}
\counterwithin{table}{section}
\renewcommand\thetable{\thesection.\arabic{table}}
\setcounter{figure}{0}
\counterwithin{figure}{section}
\renewcommand\thefigure{\thesection.\arabic{figure}}
Figure \ref{fig:region} shows the distribution of post origins. A word cloud of the text data of the unlabeled dataset collected in Section \ref{sec:trend}, and also the frequency distribution of the posts by word count, are shown in Figure \ref{fig:postinsights}.

\begin{figure*}[htbp]
    \centering
    \begin{subfigure}[m]{0.45\textwidth}
        \centering
        \includegraphics[width=\textwidth]{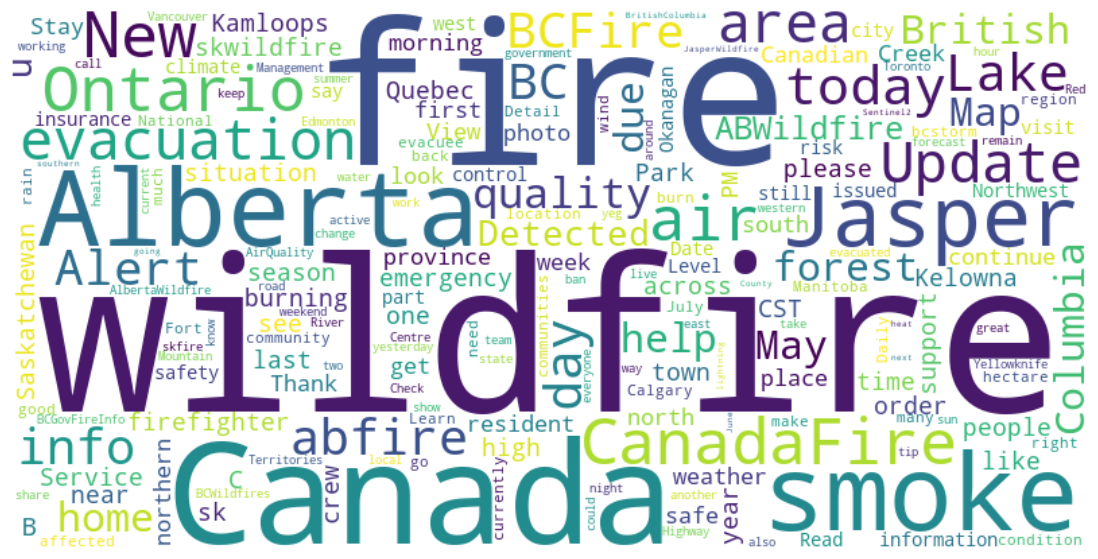}
        \caption{Word cloud of the posts.}
        \label{fig:wordcloud}
    \end{subfigure}
    \begin{subfigure}[m]{0.54\textwidth}
        \centering
        \includegraphics[width=\textwidth]{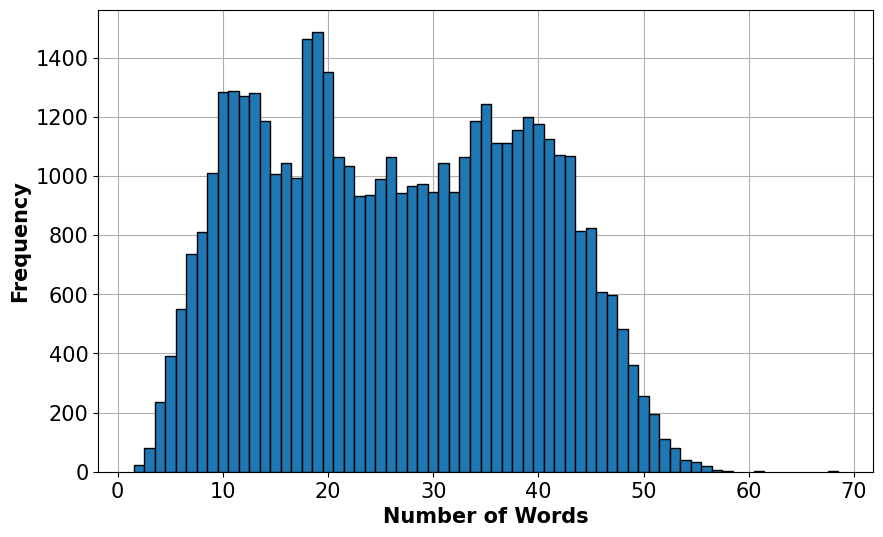}
        \caption{Frquency distribution of the posts w.r.t their lengths.}
        \label{fig:wordcount}
    \end{subfigure}
    \caption{(a) Word cloud of the text from the unlabeled tweets and (b) the frequency distribution of their word counts.}
    \label{fig:postinsights}
\end{figure*}

\end{document}